\documentclass[lettersize,journal]{IEEEtran}
\usepackage{amsmath,amssymb,amsfonts}
\usepackage{algorithmic}
\usepackage{amsmath}
\usepackage{graphicx}
\usepackage{multirow}
\usepackage{textcomp}
\usepackage{booktabs}
\usepackage{makecell}
\usepackage{gensymb}
\usepackage{bm}
\usepackage{ulem}
\usepackage{multirow}
\usepackage{amssymb}
\usepackage{graphicx}
\usepackage{xcolor}
\def\BibTeX{{\rm B\kern-.05em{\sc i\kern-.025em b}\kern-.08em
    T\kern-.1667em\lower.7ex\hbox{E}\kern-.125emX}}
\usepackage{balance}
\begin{document}
\title{Text-driven 3D Human Generation via Contrastive Preference Optimization}
\author{Pengfei Zhou, Xukun Shen, Yong Hu
\thanks{Manuscript created March, 2025; 	

Pengfei Zhou, Xukun Shen, Yong Hu are with the State Key Laboratory of Virtual Reality Technology and Systems, Beihang University, Beijing 100191, China(email: pengfeiz96@163.com; xkshen@buaa.edu.cn; huyong@buaa.edu.cn).	

This work is distributed under the \LaTeX \ Project Public License (LPPL) ( http://www.latex-project.org/ ) version 1.3. A copy of the LPPL, version 1.3, is included in the base \LaTeX \ documentation of all distributions of \LaTeX \ released 2003/12/01 or later. The opinions expressed here are entirely that of the author. No warranty is expressed or implied. User assumes all risk.}}

\markboth{Journal of \LaTeX\ Class Files,~Vol.~18, No.~9, September~2020}%
{How to Use the IEEEtran \LaTeX \ Templates}

\maketitle

\begin{abstract}

Recent advancements in Score Distillation Sampling (SDS) have significantly improved text-driven 3D human generation, enabling plausible synthesis from natural language descriptions. However, existing methods struggle with fine-grained semantic alignment, leading to inconsistencies in structure, texture, and attribute representation. To address these challenges, we propose a contrastive preference modeling framework that integrates both positive and negative prompts to enhance SDS. This approach refines text-conditioned generation, improving semantic consistency and 3D fidelity.
Specifically, we introduce a preference optimization module that aggregates knowledge from multiple preference models to ensure accurate representation of intricate attributes and nuanced details. Additionally, we propose a negation preference module with static-dynamic negation prompts to systematically distinguish relevant details from extraneous elements, ensuring geometric and textural consistency. To further mitigate artifacts and prevent reward hacking, we develop an LLM-driven dynamic negation prompting mechanism that analyzes textual inputs, decomposes complex attributes, and filters irrelevant components.
Extensive experiments demonstrate that our method achieves state-of-the-art performance, significantly enhancing texture realism, structure coherence, and overall visual alignment, particularly for complex and lengthy textual descriptions.

\end{abstract}


\begin{figure}[!htbp]
	\centering{
		\includegraphics[width=1.0\columnwidth]{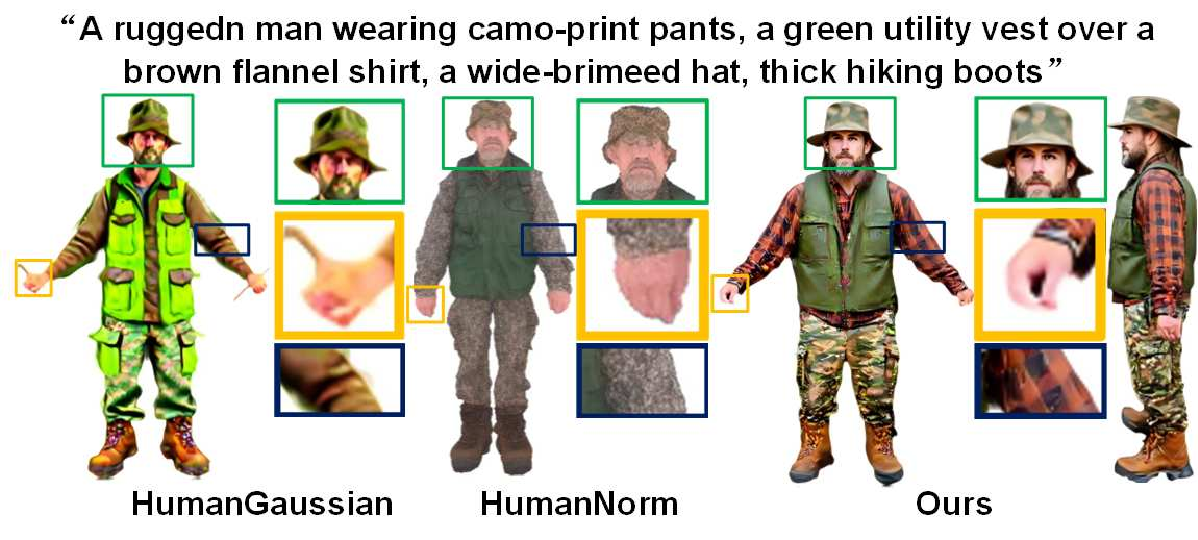}
		\caption{Comparison of Long-Text 3D Human Generation Methods. We compare the state-of-the-art methods, HumanGaussian \cite{humangaussian} and HumanNorm \cite{humannorm}, for generating 3D humans from complex long-text descriptions. The results show that these methods struggle to achieve full semantic alignment, with elements such as the green vest and brown shirt being partially occluded or incorrectly altered by other colors. This misalignment stems from the semantic entanglement effect induced by diffusion priors, which disrupts fine-grained attribute mapping and leads to incomplete text-to-3D correspondence in long-text-conditioned generation.}
		\label{fig2:over}
	}
\end{figure}

\section{Introduction}
\IEEEPARstart{T}{he} creation of high-fidelity 3D human models from detailed textual descriptions has attracted widespread attention in the industry, with applications spanning VR/AR, filmmaking, animation production, the metaverse, and more. Traditional methods for customizing personalized avatars are time-consuming and labor-intensive. To automate 3D asset generation, generative models such as Generative Adversarial Networks (GANs) and diffusion models have been integrated into 3D pipelines. However, the scarcity of large-scale 3D datasets and the high computational cost of training remain critical challenges. To address these issues, DreamFusion \cite{dreamfusion} introduces score distillation sampling (SDS) to leverage 2D prior knowledge from images for distilling 3D representations. This method produces reasonable results for single objects but struggles to model complex human bodies, resulting in problems such as Janus artifacts, geometric inconsistencies, and blurriness.

Several approaches \cite{tada,dreamwaltz,humangaussian} combine SDS with the SMPL-X \cite{smplx} model to mitigate Janus artifacts and geometric inconsistencies. For example, TADA \cite{tada} proposes a subdivided SMPL-X to replace implicit representations, achieving high-quality 3D human generation. DreamWaltz \cite{dreamwaltz} utilizes ControNet \cite{contronet} and SMPL-X to provide skeletal viewpoint-consistent 2D human guidance to resolve 3D geometric consistency issues. HumanGaussin \cite{humangaussian} uses 3D Gaussian Splatting \cite{3dgs} (3DGS) and SMPL-X to represent 3D humans, collecting dense points from SMPL-X as initial point clouds for generating high-quality 3D human models. Additionally, HumanNorm \cite{humannorm} employs high-resolution tetrahedral human priors and fine-tuned diffusion models to refine normal and depth map estimation, improving geometric and texture consistency.

However, these methods continue to exhibit representation inconsistencies, resulting in misalignments in geometry, texture, and attributes when processing textual descriptions. The primary issue stems from the implicit semantic entanglement of SDS priors. Specifically, when textual prompts are fed into the diffusion model, it retrieves the closest semantic approximation rather than achieving precise textual alignment, leading to geometric distortions and texture deviations. This issue becomes particularly pronounced in complex descriptions, as shown in Fig. \ref{fig2:over}. HumanGaussian \cite{humangaussian} exhibits geometric distortions in hand generation, while HumanNorm \cite{humannorm} produces structural artifacts in generated accessories, highlighting deficiencies in geometric reasoning. Furthermore, both methods suffer from texture leakage, such as unnatural neon hue shifts in clothing regions and misinterpretation of fabric patterns, revealing inconsistencies in texture preservation and attribute mapping. To address these limitations, we introduce a human preference-aware optimization framework to enhance text-to-3D alignment. By leveraging human-level preference modeling, our method improves attribute perception and strengthens semantic control, effectively preserving fine-grained details from textual descriptions. This ensures more faithful and consistent 3D human generation, particularly for highly complex prompts. 

In this paper, we propose a novel framework to address the limitations of existing methods in fine-grained semantic alignment when processing long and complex textual descriptions. At the core of our approach is the introduction of contrastive preferences, which leverage human-level preference models to enhance text-conditioned 3D generation. By incorporating both positive and negative prompts, our method refines the semantic understanding of textual inputs, improving the fidelity, consistency, and controllability of the generated 3D humans.  

Specifically, we propose a preference optimization module that integrates multiple preference models, enabling the framework to effectively extract and disentangle diverse semantic information from lengthy textual descriptions. Through fine-grained attribute modeling, this module mitigates the semantic entanglement effect induced by diffusion priors in SDS, enhances the perception of comprehensive textual semantics, and ensures consistency in geometry, texture, and attribute mapping for the generated 3D human models.
To balance the contributions of different preference models, we introduce an adaptive preference weight fusion strategy. This strategy employs a least common multiple (LCM)-based dynamic weighting mechanism, which iteratively adjusts preference weights based on individual model scores and their computed LCM. By dynamically calibrating the influence of each preference model, our approach effectively leverages their complementary semantic perception capabilities, significantly improving the fidelity of complex attributes and fine-grained details.

Furthermore, we propose a negation preference optimization module to address reward hacking in preference models by integrating negation prompts as additional constraints. This module systematically filtering out extraneous information while preserving essential semantic details. This ensures greater consistency in geometry, texture, and attribute mapping, leading to more faithful and visually coherent text-to-3D human generation.
Our module integrates both static and dynamic negation prompts to ensure the comprehensiveness of the negation constraints. To further enhance adaptability, we introduce a large language model (LLM)-driven dynamic negation prompting mechanism, which dynamically analyzes textual inputs, decomposes fine-grained clothing attributes, and systematically recombines adjective-attribute pairs (e.g., “white canvas shoes, red jacket” → “red canvas shoes, white jacket”). Additionally, the LLM estimates potentially irrelevant elements (e.g., “baseball cap” → “baseball glove”), proactively reducing unintended attribute entanglement. This adaptive negation strategy prevents over-optimization of dominant features while preserving a balanced and faithful semantic representation, thereby enhancing the realism and controllability of text-driven 3D human generation.

Our contributions are summarized as follows:
\begin{itemize}
	\item 
	 Firstly, we propose a preference optimization module that incorporates an adaptive weight fusion strategy to achieve complementary semantic perception. Our method effectively mitigates the semantic entanglement effect induced by diffusion priors, thereby enhancing the alignment of complex attributes and fine-grained details in text-to-3D human generation.
	\item 
    Furthermore, we propose a negation preference optimization module that integrates both static and LLM-driven dynamic negation prompts to systematically regulate attribute mapping and refine semantic control. By dynamically analyzing textual descriptions, decomposing and recombining adjective-attribute pairs, and filtering irrelevant elements, our approach effectively mitigates attribute entanglement and reward hacking, ensuring a more balanced, semantically consistent, and realistic 3D human representation.
	\item 
	Finally, extensive quantitative experiments demonstrate that our method outperforms recent approaches in terms of textual alignment accuracy. Qualitative results further show that our method achieves superior visual quality, generating more semantically accurate and visually coherent 3D models compared to existing methods.
\end{itemize}

\section{RELATED WORK}\label{2}
\subsection{Text-to-3D Generation}\label{2.1}
With the rapid advancements in text-to-image generation models, researchers increasingly focus on extending these capabilities to text-to-3D generation \cite{magic3d,prod, mvdream,mvgaussian}. Early approaches \cite{avatarclip,clipmesh,clipforge} leverage CLIP-based loss functions to supervise 3D representation learning, achieving promising results. However, these methods often struggle with realism and fine-grained geometric details. A significant breakthrough comes with DreamFusion\cite{dreamfusion}, which replaces CLIP supervision with a text-to-image diffusion model and introduces SDS to optimize NeRF-based 3D representations. This innovation lays the foundation for numerous follow-up methods.
Magic3D \cite{magic3d} improves computational efficiency through a two-stage pipeline, incorporating a more effective DMET \cite{dmtet} representation. ProlificDreamer \cite{prod} addresses over-saturation and over-smoothing issues by introducing a variational SDS objective, while MVDream \cite{mvdream} employs a multi-view diffusion model to enhance 3D consistency across different viewpoints. Fantasia3D \cite{fantasia3d} introduces a decoupled geometry and appearance modeling strategy, optimizing surface normals independently using SDS loss. More recently, DreamGaussian \cite{dreamgaussian} replaces NeRF with 3DGS, significantly reducing inference time. MVGaussian \cite{mvgaussian} further refines this approach by introducing a densification algorithm that aligns Gaussian distributions with the surface, thereby improving structural fidelity.

Further advancements target structural coherence and optimization efficiency. DreamCouple \cite{dreamcouple} introduces a correction flow model to identify coupled noise and incorporates a unique pair matching (UCM) loss to mitigate over-smoothing effects. DreamMapping \cite{dreammapping} proposes Variational Distribution Mapping (VDM) and timestep-dependent Distribution Coefficient Annealing (DCA) to enhance distillation precision. ISD refines the SDS framework by replacing its reconstruction term with an invariant score term derived from DDIM sampling, enabling a moderate classifier-free guidance scale. This modification reduces reconstruction errors while alleviating over-smoothing and over-saturation issues.

Despite these advances, current methods still face significant challenges in high-quality 3D human generation, particularly in mitigating Janus artifacts and geometric inconsistencies, highlighting the need for further refinement in fine-grained semantic alignment and structural fidelity.
\subsection{Text-driven 3D Human Generation}\label{2.2}
Existing methods for text-driven 3D human generation leverage various representations and optimization strategies, yet they still face challenges in realism, detail preservation, and alignment with complex textual descriptions. AvatarCLIP \cite{avatarclip} integrates SMPL \cite{spml} and NeuS \cite{neus} with CLIP supervision to generate 3D avatars, but the results often lack realism and appear overly simplified. DreamAvatar \cite{dreamavatar} and AvatarCraft \cite{avatarcraft} incorporate SMPL priors for pose and shape constraints, yet they tend to produce blurry 3D avatars. DreamWaltz \cite{dreamwaltz} enhances generation quality by utilizing 3D-consistent occlusion-aware SDS with 3D-aware skeletal conditions, while DreamWaltz-G \cite{dreamwaltzg} further refines this approach by introducing a hybrid 3D Gaussian avatar representation, combining neural implicit fields and parametric 3D meshes to enable real-time rendering and stable SDS optimization.

Other methods focus on pose conditioning and structured diffusion models. DreamHuman \cite{dreamhuman} generates animatable 3D humans by incorporating pose-conditioned NeRF trained on imGHUM \cite{imghum}. AvatarBooth \cite{avatarbooth} employs dual fine-tuning diffusion models to separately refine facial and body details, improving the personalization of avatars from casual images. AvatarVerse \cite{avatarverse} enhances view consistency by training ControlNet \cite{contronet} conditioned on DensePose \cite{densepose}. TADA \cite{tada} utilizes SMPL-X with displacement layers and texture maps, introducing hierarchical rendering with SDS loss to improve 3D human synthesis. Building upon this framework, X-Oscar \cite{xoscar} proposes avatar-aware score distillation sampling (ASDS), which injects avatar-specific noise into rendered images to enhance optimization and improve generation quality. GAvatar \cite{gavatar} introduces an SDF-based implicit mesh learning approach, capturing fine facial geometries and enabling large-scale animatable avatar synthesis from textual descriptions.

More recently, fine-tuned text-to-image methods have been explored to enhance text-driven 3D human generation. HumanNorm \cite{humannorm} fine-tunes a text-to-image model to predict normal and depth maps from text, which are subsequently used for SDS-based geometry refinement and texture optimization via normal-conditioned ControlNet \cite{contronet}. HumanGaussian \cite{humangaussian} introduces structure-aware SDS, jointly optimizing geometry and appearance by fine-tuning a text-to-image model to predict depth maps, which guide Gaussian densification and pruning to enhance human body synthesis. While these methods improve the structural and appearance quality of generated avatars, they struggle with semantic alignment for long textual descriptions, limiting their ability to capture complex, fine-grained attributes.
To overcome these limitations, we propose a novel framework that enhances long-text alignment in text-to-3D human generation, ensuring more precise and semantically consistent 3D outputs.

\begin{figure*}[!htbp]
	\centering{
		\includegraphics[width=2.0\columnwidth]{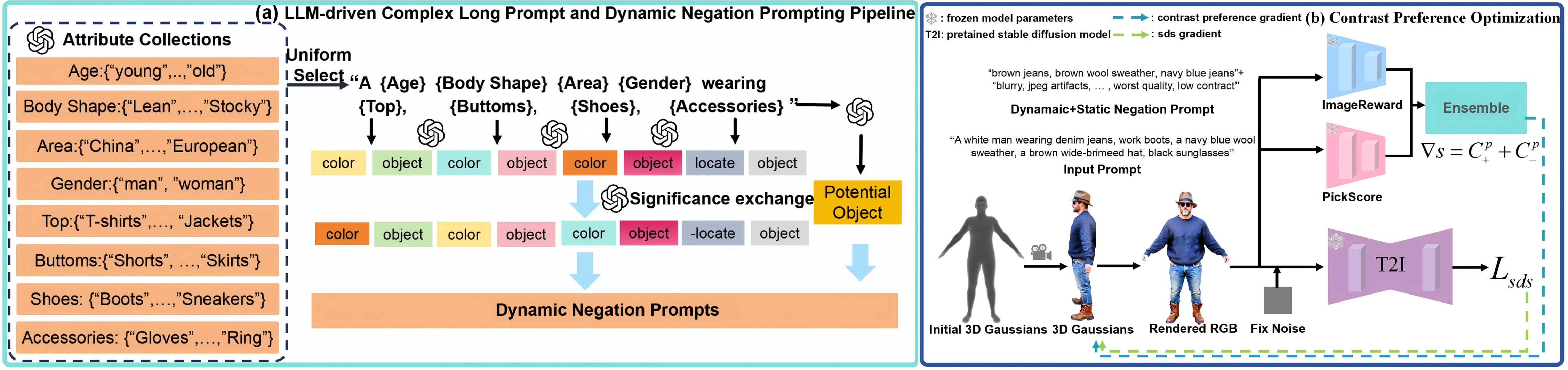}
		\caption{Overview of the Proposed Framework. We first leverage a LLM to generate complex long-text prompts along with dynamic negative prompts, which guide the optimization process and enhance the quality of 3D human generation. Next, we employ the representation of 3DGS \cite{3dgs} to synthesize accurate 3D human models from the generated textual descriptions. The process begins with densely sampling Gaussian functions on the SMPL-X \cite{smplx} human mesh, serving as the initialization for model generation. To improve semantic alignment between the long-text descriptions and the 3D human models, we introduce a preference optimization module, which integrates preference models such as ImageReward \cite{imagereward} and PickScore \cite{pickscore}. These models capture fine-grained semantic nuances across different text segments, enhancing overall alignment accuracy. Additionally, we propose a negation preference optimization module, which incorporates both static and dynamic negative prompts to mitigate over-optimization in preference models. This mechanism prevents excessive emphasis on certain attributes while ensuring balanced semantic representation. The entire framework is optimized through SDS, where preference models refine the SDS process to achieve more precise alignment between long-text descriptions and 3D human representations.}
		\label{fig:pipelines}
	}
\end{figure*}

\subsection{Text-to-Image Generative Models Alignment}\label{2.3}
Recent advancements in text-to-image generation have significantly improved the quality of synthesized 2D images. However, these images often fail to align precisely with human preferences, making preference alignment a critical research challenge. To address this issue, various studies \cite{imagereward,pickscore,visionprefer,longalign} have focused on constructing large-scale text-image preference datasets and fine-tuning visual models as evaluation functions, thereby enhancing the alignment between generated outputs and human expectations.

Several notable approaches have been proposed to improve preference alignment. ImageReward \cite{imagereward} introduces a reward-feedback learning framework to optimize diffusion models and presents the first universal text-image human preference reward model, effectively encoding human visual preferences. PickScore \cite{pickscore} constructs a large-scale open dataset for text-to-image generation and employs a fine-tuned CLIP model to enhance preference alignment. HPS\_V2 \cite{hps} curates a high-quality preference dataset with carefully selected text-image pairs to mitigate potential biases. VisionPrefer \cite{visionprefer} utilizes a multimodal large-scale language model to generate a fine-grained preference dataset, incorporating key factors such as prompt adherence, aesthetics, fidelity, and harmlessness to enhance visual preference modeling. To further improve alignment in long-text-driven generation, LongAlign \cite{longalign} introduces a segment-level encoding approach, facilitating the processing of complex textual descriptions and leveraging a CLIP-based DenseScore preference model for efficient long-text alignment training.

Despite these advancements, existing preference models still exhibit semantic bias issues. For instance, ImageReward effectively recognizes categorical attributes (e.g., color and position) but struggles to capture fine-grained textual attributes (e.g., specific hat types). In contrast, PickScore demonstrates superior attribute perception, accurately identifying fine-grained object properties such as different types of hats. To address these limitations, we propose an integrated preference modeling approach that combines multiple preference models, enabling a more comprehensive understanding of textual semantics and improving the robustness and accuracy of preference alignment.

\section{Methodology}\label{3}

An overview of our method is illustrated in Fig. \ref{fig:pipelines}. Our proposed pipeline consists of four key components: First, we leverage 3DGS \cite{3dgs} and SMPL-X \cite{smplx} to represent 3D human bodies and introduce an optimization paradigm for refining the 3D Gaussian representation through SDS optimization, as detailed in Sec. \ref{3.1}.
Additional, we propose an LLM-driven complex long prompt and dynamic negation prompt construction module. This module utilizes an LLM to generate complex prompts and dynamic negation prompts based on a generation template and ranking rules. By integrating positive and dynamic negation prompts, this module enhances semantic understanding of textual input and improves the fidelity, consistency, and controllability of the generated 3D human models, as elaborated in Sec. \ref{3.2}.
Furthermore, we introduce the preference optimization module, which leverages multiple preference models to capture semantic attributes from the text and generate fine-grained semantic details in the 3D human body. The enhancement in semantic alignment is discussed in Sec. \ref{3.3}.
Finally, we present the negation preference optimization module, which integrates additional static-dynamic hybrid negation prompts to mitigate the generation of irrelevant objects caused by preference model over-optimization and improve texture clarity, as described in Sec. \ref{3.4}.

\subsection{Preliminaries}\label{3.1}
\noindent \textbf{SMPL-X} \cite{smplx} is a 3D parametric model used for modeling human body shape, encompassing the topology of the body, hands, and face. The model consists of 10,475 vertices and 54 keypoints. By combining pose parameters $\theta'$ (which include body pose ${{\theta' }_{\text{b}}}$, facial pose ${{\theta' }_{\text{f}}}$, and finger pose ${{\theta'}_{\text{h}}}$), shape parameters $\beta$, and expression parameters $\Psi $, we can represent the 3D SMPL-X human model as $M(\beta ,\theta' ,\Psi )$:
\begin{align}
	& T(\beta ,\theta' ,\Psi )=\bar{T}+{{B}_{s}}(\beta )+{{B}_{p}}(\theta' )+{{B}_{e}}(\Psi ), \\ 
	& M(\beta ,\theta' ,\Psi )=LBS(T(\beta ,\theta' ,\Psi ),J(\beta ),\theta' ,\omega ), 
\end{align}
where $\bar{T}$ represents the mean template shape; ${B_s}$, ${B_p}$ and ${B_e}$ are the blending shape functions for shape, pose, and expression, respectively. $T(\beta ,\theta' ,\Psi )$ is a non-rigid deformation starting from $\bar{T}$. The function $LBS$ \cite{lbs} is the linear blend skinning function, which transforms $T(\beta ,\theta' ,\Psi )$ into the target pose $\theta'$. It defines the skeletal joints $J(\beta)$ and blending weights $\omega$ for each vertex.

\noindent \textbf{Score Distillation Sampling} extracts the 2D pre-trained diffusion prior for optimizing the 3D representation, a 3D scene with $\theta$ and use a differentiable rendering function $g(.)$ to obtain an image $x=g(\theta)$. By pushing the samples towards the denser regions of the real data distribution across all noise levels, ensuring that the rendering from each camera view closely resembles reasonable samples $\phi$ derived from the guidance diffusion model. In practice, DreamFusion \cite{dreamfusion} employs Imagen \cite{imagen} as the score estimation function ${{\varepsilon }_{\phi }}({{x}_{t}},y)$, which predicts the sampling noise given a noisy image ${x_t}$, a text embedding $y$, and a time step $t$. SDS then optimizes the 3D scene using gradient descent on $\theta$:
\begin{equation}
{{\nabla }_{\theta }}{{L}_{sds}}={{E}_{\varepsilon ,t}}[{{\omega }_{t}}({{\varepsilon }_{\phi }}({{x}_{t}};y)-\varepsilon )\frac{\partial x}{\partial \theta }].
\end{equation}

\noindent \textbf{3D Gaussian Splatting} \cite{3dgs}: provides an efficient representation for novel view synthesis and 3D reconstruction. Unlike implicit methods such as NeRF \cite{nerf}, 3D Gaussian Splatting represents the underlying scene using a set of anisotropic Gaussian functions. The parameters of these functions include the center position $\mu \in {{\mathbb{R}}^{3}}$, covariance $\sum \in {{\mathbb{R}}^{7}}$, color $c\in {{\mathbb{R}}^{3}}$, and opacity $\alpha \in \mathbb{R}$. By projecting the 3D Gaussian functions onto the camera's imaging plane, 2D Gaussian functions are assigned to corresponding tiles for point-based rendering:
\begin{align}
	& G(p,{{\mu }_{i}},{{\Sigma }_{i}})=\exp (-\frac{1}{2}(p-{{\mu }_{i}})), \\ 
	& c(p)=\sum\limits_{i\in N}{{{c}_{i}}{{\sigma }_{i}}\underset{j=1}{\overset{i-1}{\mathop{\prod }}}\,(1-{{\sigma }_{j}}),{{\sigma }_{i}}={{\alpha }_{i}}G(p,{{\mu }_{i}},{{\Sigma }_{i}}),}  
\end{align}
where $p$ represents the query point's position; ${{\mu}_{i}}$, ${{\Sigma}_{i}}$, ${c_i}$, ${{\alpha}_{i}}$ and ${{\sigma}_{i}} $, are the center position, covariance, color, opacity, and density of the i-th Gaussian distribution, respectively. ${ G(p,{{\mu }_{i}},{{\Sigma }_{i}})}$ is the value of the i-th Gaussian function at point $p$. $N$ denotes the set of 3D Gaussians within the tile. 

\subsection{LLM-driven Complex Long Prompt and Dynamic Negation Prompting Pipeline} \label{3.2}


To effectively validate the proposed method for generating high-fidelity 3D human models with comprehensive semantic alignment from lengthy and complex textual descriptions, we design a 3D human generation template, as illustrated in Fig. \ref{fig:pipelines}-(a). This template encompasses basic human attributes such as age, body shape, region, and gender, along with apparel attributes including upper clothing, lower clothing, shoes, and accessories. Notably, age, body shape, and region are randomly selected to ensure diversity. (The design avoids any skin color discrimination.) Additionally, all generated textual descriptions are unique. Following this template, we utilize a LLM to generate 1,000 diverse and complex human-centric prompts, from which 100 prompts are randomly sampled for evaluation in our experiments.

To further mitigate semantic entanglement caused by diffusion priors and enhance the controllability of 3D human generation, we introduce an LLM-driven dynamic negative prompt strategy. Specifically, we first parse the descriptive text to extract modifier-attribute pairs (MAPs) related to clothing and accessories. The LLM then evaluates the saliency of each modifier (e.g., "bright colors" are more visually dominant) and reconstructs the MAPs based on saliency, forming negated MAPs (e.g., "red canvas shoes" → "white canvas shoes").

For attributes involving spatial positioning (e.g., "black glove on the left hand"), we reverse the position (e.g., "black glove on the right hand") to prevent positional overfitting. Furthermore, to suppress irrelevant objects introduced by diffusion priors and preference models, the LLM is leveraged to identify potential irrelevant elements (e.g., "baseball cap" → "baseball glove").

By integrating the reordered MAPs and irrelevant objects, our dynamic negative prompting strategy effectively prevents over-optimization of dominant features while maintaining semantic fidelity and balance, thereby significantly enhancing the realism and controllability of text-driven 3D human generation.

\subsection{Preference Optimization Module (POM)}\label{3.3}

To mitigate the semantic entanglement effect induced by diffusion priors in SDS, we introduce a human-level preference-driven module that effectively captures fine-grained semantics from long and complex textual descriptions. Unlike prior methods that rely on a single preference model, our approach leverages multiple complementary preference models to enhance the geometric accuracy and texture fidelity of the generated 3D human models.

\begin{figure}[!htbp]
	\centering{
		\includegraphics[width=1.0\columnwidth]{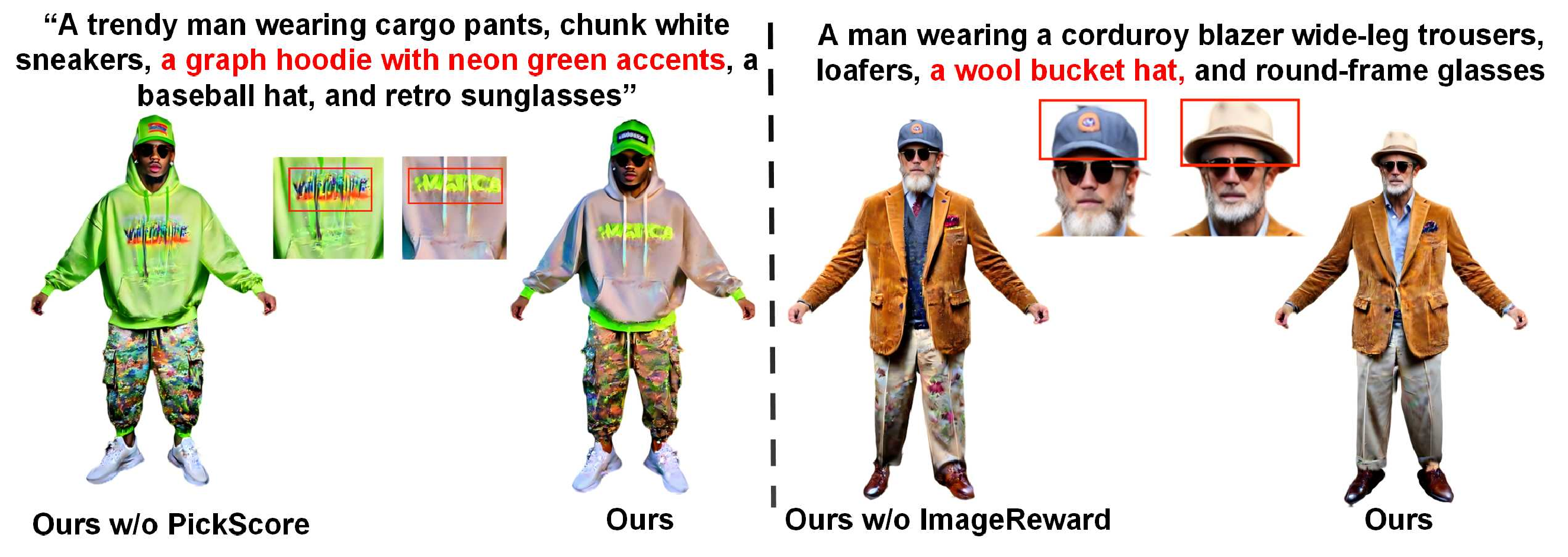}
		\caption{PickScore \cite{pickscore} demonstrates a stronger capability in capturing categorical semantics, accurately generating a bucket hat, whereas ImageReward \cite{imagereward} fails to distinguish the category precisely, resulting in a baseball cap. In contrast, ImageReward excels in attribute-level semantic understanding, correctly identifying the pattern's color, while PickScore misinterprets it as the hoodie’s color, leading to an incorrect generation.}
		\label{fig77}
	}
\end{figure}

Specifically, we observe that single preference models exhibit inherent semantic bias. For instance, ImageReward (IR) demonstrates superior sensitivity to local visual attributes, such as color and texture, while PickScore (PS) excels in object-level recognition, such as accessory categories like hats and glasses. As shown in Fig. \ref{fig77}, the IR effectively restores text-guided color semantics, whereas PS accurately captures semantic details related to object categories.
By integrating these complementary signals, our module achieves fine-grained semantic alignment, leading to significant improvements in geometric consistency and texture realism.

Moreover, we adopt an adaptive weighting strategy to dynamically balance the contributions of each preference model, further enhancing the overall quality of 3D human generation.
Specifically, we feed the rendered 3D human images and long textual descriptions into multiple preference models to obtain the corresponding preference scores. Due to the inherent differences in semantic perception across models, directly adopting equal-weighted optimization often leads to semantic misalignment and loss of fine-grained details. To address this issue, we propose a Least Common Multiple (LCM) based Adaptive Weighting Strategy.

Given the preference models \(P = \{IR, PS\}\), the long text description \(Y = \{y_1, y_2, \dots, y_k\}\), and the rendered image \(X\), we first apply the proposed module to obtain the preference scores and gradients:

\begin{equation}
	s_{IR}, \nabla\ s_{IR}, s_{PS}, \nabla s_{PS} = \{IR(X, Y), PS(X, Y)\}.
\end{equation}
Next, we compute the LCM of all preference scores LCM(${{s}_{IR}}$, ${{s} _{PS}}$).
Then, an inverse weighting strategy is applied, where models with lower preference scores (i.e., those that are under-optimized in the current stage) are assigned higher weights to boost their contribution during the optimization process:

\begin{equation}
  {\lambda}_i = \sum\limits_{i=1}^{N=2} \frac{\text{LCM}(s_{IR}, s_{PS})}{s_i}.
\end{equation}

Finally, the semantic information integrated through this LCM-based adaptive weighting strategy significantly enhances geometric consistency, texture details, and semantic alignment in 3D human generation. The final preference function can be formulated as:

\begin{equation}
	C_{+}^{p}=\frac{1}{N}\sum\limits_{i={1}}^{N={2}}{{{\lambda }_{i}}{\nabla s_i}},
\end{equation}
where $N$ is the number of preference models, and $\lambda_i$ represents the weight of the i-th preference model.

\subsection{Negative Preference Optimization Module (NPOM)}\label{3.4}


To mitigate attribute entanglement and prevent reward hacking from deviating the optimization process from the target direction, we propose a comprehensive negation prompting strategy to guide the optimization process, ensuring more balanced, semantically consistent, and realistic 3D human generation.

Our comprehensive negation prompt consists of static and dynamic negation prompts. The static negation prompt is a fixed set of negative phrases, similar to the negative prompts used in diffusion models. We define the static negation prompt as:
static\_neg\_prompt = \{`blurry, oversaturated, noisy, ..., jpeg artifacts'\}.
The static negation effectively alleviates issues such as artifacts, blurriness, and noise.

To further guide the optimization direction, we introduce an LLM-driven dynamic negation prompt. By dynamically analyzing the input text, the LLM decomposes fine-grained adjective-attribute pairs and reorders them based on attribute saliency (e.g., “white canvas shoes, red jacket” → “red jacket, white canvas shoes”). This allows the optimization process to prioritize more visually dominant attributes. Additionally, the LLM estimates potential irrelevant elements (e.g., “baseball cap” → “baseball glove”) and proactively suppresses unintended attribute entanglement.

By integrating static and dynamic negation prompts, our method effectively steers the optimization process towards the target direction, significantly enhancing the realism, consistency, and controllability of text-driven 3D human generation.

Given the negation prompt Y\_neg, we apply the proposed model to obtain the negation preference scores and gradients {IR(X, Y\_neg), PS(X, Y\_neg)}:

\begin{equation}
	s_{IR}, \nabla s_{IR}, s_{PS}, \nabla s_{PS} = \{IR(X, Y_{neg}), PS(X, Y_{neg})\}.
\end{equation}

Subsequently, we use the averaged negation preference scores to constrain the preference model and prevent overfitting. The negation preference function can be defined as:
\begin{equation}
	C_{-}^{p}=\frac{1}{N}\sum\limits_{i=1}^{N=2}{{{\nabla s_i}}}.
\end{equation}

The complete contrastive preference function can be defined as:
\begin{equation}
	C_{all}^{p}=C_{+}^{p}+C_{-}^{p}.
\end{equation}

In summary, the objective function can be defined as:

\begin{equation}
    {{L}_{all}}=C_{all}^{p}+{{L}_{sds}}.
\end{equation}

\section{Experiments}\label{4}

\begin{figure*}[!htbp]
 	\centering{
 		\includegraphics[width=2.0\columnwidth]{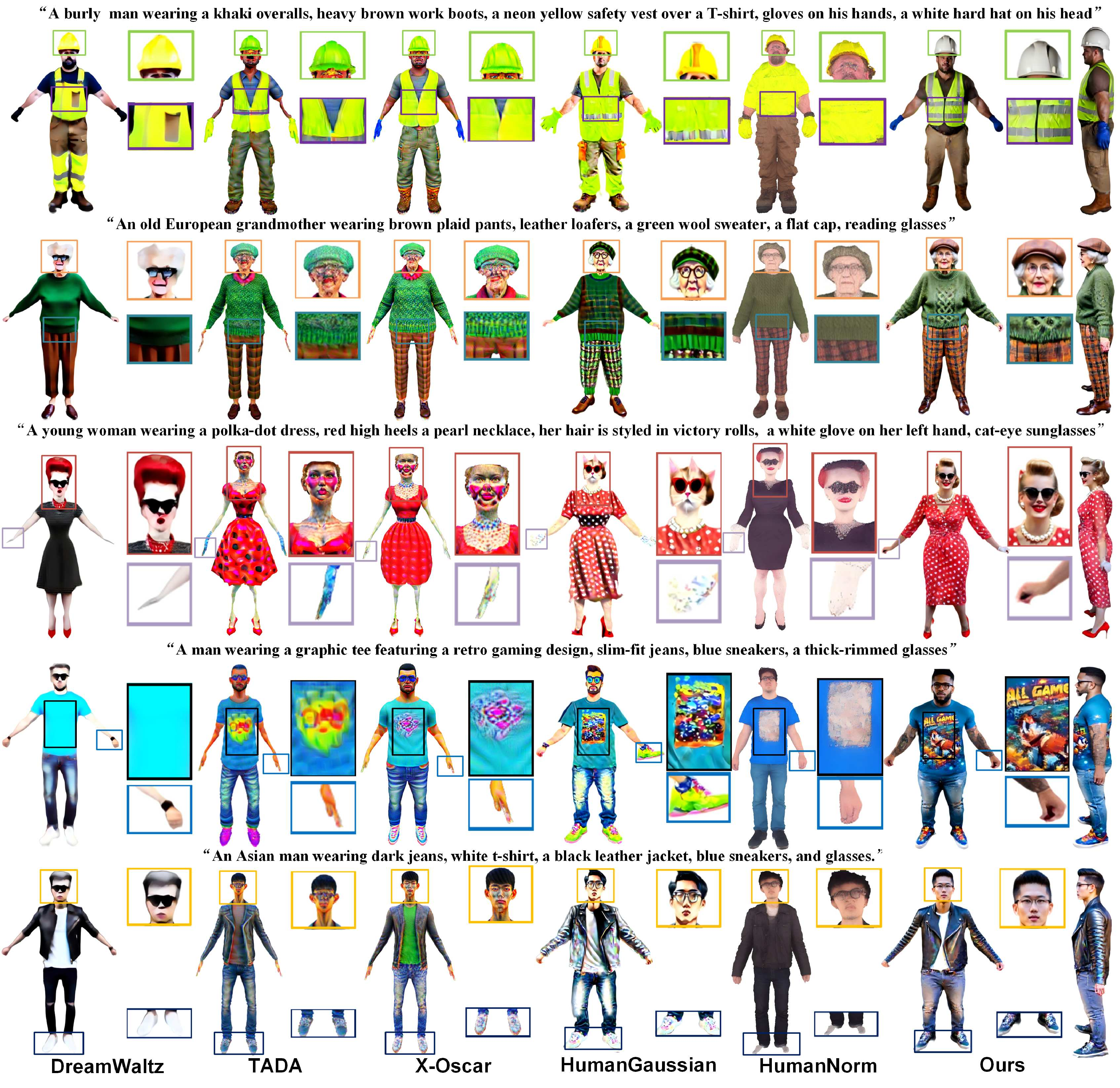}
 		\caption{Qualitative comparisons with the SOTA Text-to-3D human methods. Compared to existing text-driven 3D human generation methods, such as DreamWaltz \cite{dreamwaltz}, TADA \cite{tada}, X-Oscar \cite{xoscar}, HumanNorm \cite{humannorm}, and HumanGaussian \cite{humangaussian}, our approach demonstrates significant improvements. The text prompts used for comparison are listed at the top.}
 		\label{fig}
 	}
\end{figure*}

\subsection{Experiment Setups}\label{4.1}
\noindent \textbf{Training and 3D GS Setups.} 
The 3D Gaussian model is initialized with 100k instances, uniformly sampled on the SMPL-X mesh surface with an opacity of 0.1. The entire 3DGS training process spans 3600 iterations, with dense sampling and pruning conducted every 300 steps between 300 to 2100 iterations. The pruning phase is executed with a threshold factor of 0.008, applied every 300 steps from iteration 2400 to 3300. The framework is optimized using the Adam optimizer, with $\beta$ values of [0.9, 0.99]. The learning rates for the center position $\mu$, scale factor $s$, rotation quaternion $q$, color $c$, and opacity $\alpha$ are set to $5 \times 10^{-5}$, $1 \times 10^{-3}$, $1 \times 10^{-2}$, $1.25 \times 10^{-2}$, and $1 \times 10^{-2}$, respectively.

\noindent \textbf{Implementation details.} 
Our method is implemented in PyTorch, based on the ThreeStudio framework. We employ the Adam optimizer with a learning rate of 0.001. The camera distance is constrained within the range [1.5, 2.0], the field of view (fovy) spans [40°, 70°], the elevation angle is limited to [-30°, 30°], and the azimuth angle ranges from [-180°, 180°]. The SDS loss weight is set to 1, while the guidance scale is fixed at 7.5. We uniformly sample the time steps as $t \sim U(0.02, 0.50)$. The training resolution is 1024, with a batch size of 4. The entire optimization process takes 0.8 hours on a single NVIDIA RTX 4090 GPU.

\noindent \textbf{Baselines.} 
We conduct a comprehensive comparison of our proposed text-driven 3D human generation method with state-of-the-art (SOTA) approaches. Specifically, we evaluate its performance against DreamWaltz \cite{dreamwaltz}, TADA \cite{tada}, X-Oscar \cite{xoscar}, HumanGaussian \cite{humangaussian}, and HumanNorm \cite{humannorm}. The comparison primarily focuses on geometric consistency, texture realism, and semantic alignment, particularly in handling long and complex textual descriptions of humans. Additionally, we compare our method with text-to-3D content generation approaches such as DreamGaussian \cite{dreamgaussian}, DreamAvatar \cite{dreamavatar}, and AvatarVerse \cite{avatarverse}, emphasizing texture detail and realism.

\noindent \textbf{Metrics.} We employ CLIP ViT-B/32 \cite{clip} to evaluate the alignment between textual descriptions and rendered images. However, CLIP alone struggles with fine-grained text-to-3D correspondence. To address this, we integrate DenseScore \cite{visionprefer}, an enhanced model leveraging dense feature representations and fine-tuned on long-text preference datasets, enabling more precise semantic and spatial alignment. Additionally, we incorporate HPS\_V2 \cite{hps} to assess texture quality and aesthetic fidelity, ensuring a comprehensive evaluation of visual realism.
\subsection{Qualitative Evaluations}

\noindent \textbf{Comparison with text-to-3D human methods.} We conduct a qualitative comparison with recent methods on complex long-text descriptions. As illustrated in Fig. 
\ref{fig}, DreamWaltz \cite{dreamwaltz} employs a NeRF-based \cite{nerf} representation to ensure geometric and texture consistency. However, due to the lack of strong supervisory signals, the generated textures suffer from severe blurring and artifacts. TADA \cite{tada} and X-Oscar \cite{xoscar} optimize directly on the SMPL-X \cite{smplx} model, ensuring structural completeness and temporal coherence for animated sequences. However, the limited representational capacity of SMPL-X leads to unrealistic geometric distortions and blurry textures, reducing overall realism. For instance, in the generated results, facial features and limbs exhibit significant deformation.
HumanNorm \cite{humannorm} introduces a tetrahedral representation to enhance both geometric and texture expressiveness, achieving relatively high-quality outputs. However, its generalization capability is limited, leading to severe geometric distortions in accessories, such as glasses in Fig. \ref{fig}-(3,4) and feet in Fig. \ref{fig}-(5). HumanGaussian \cite{humangaussian} incorporates 3DGS \cite{3dgs} and a dual-branch SDS mechanism to ensure geometric consistency. However, its modified NFSD suffers from semantic inconsistencies, resulting in anomalies such as a cat-like face in Fig. \ref{fig}-(3) and a hand-to-foot transformation in Fig. \ref{fig}-(4). Moreover, all these methods exhibit texture bleeding artifacts, which stem from biases in the diffusion prior.

In contrast, our method generates 3D humans with finer and more plausible geometric and texture details. Furthermore, it ensures high fidelity to the input text without omitting any clothing items or accessories, achieving superior semantic alignment with the given prompts.

\begin{figure}[!htbp]
	\centering{
		\includegraphics[width=1.0\columnwidth]{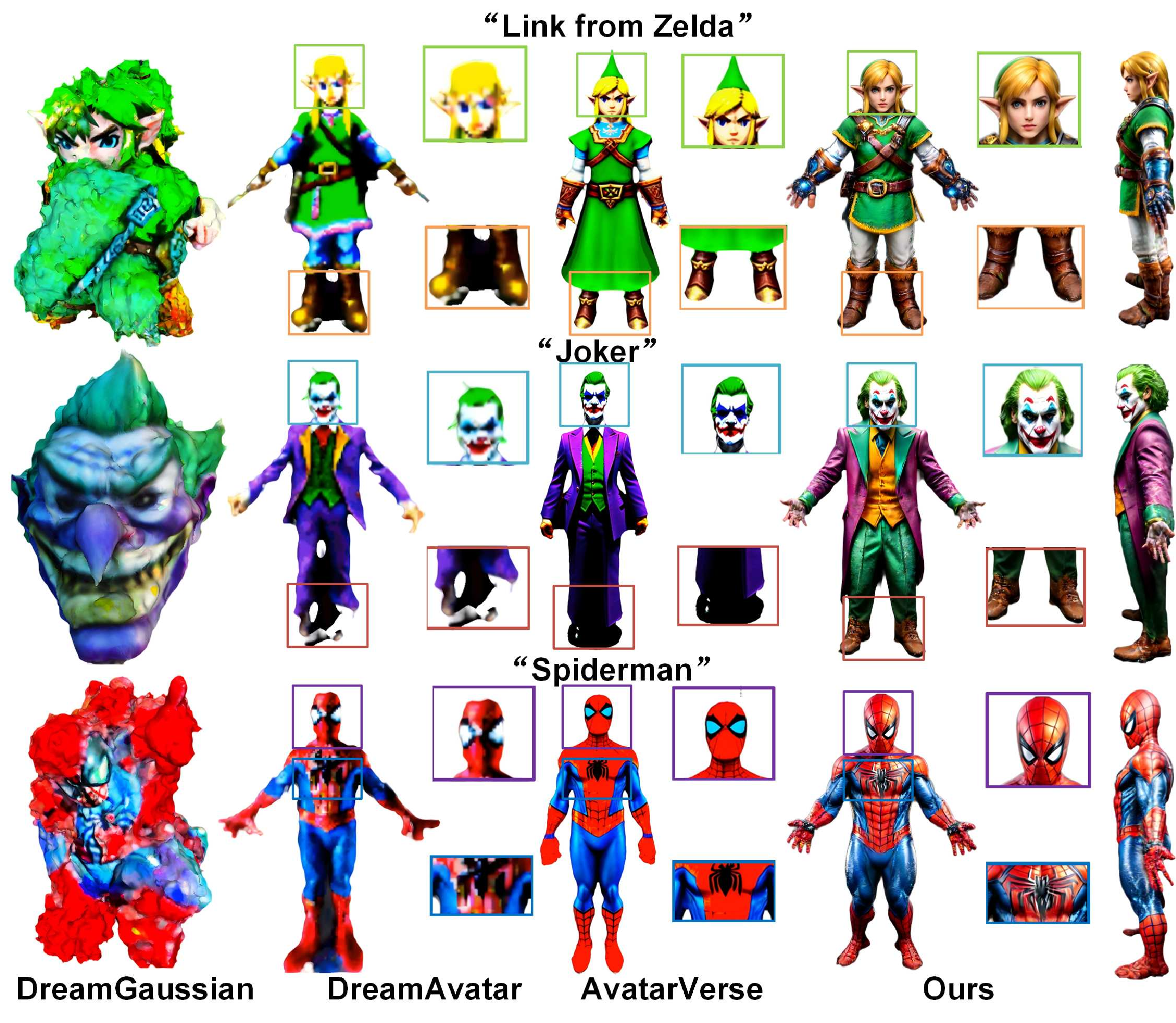}
		\caption{Qualitative comparisons with the SOTA Text-to-3D methods.}
		\label{fig5}
	}
\end{figure}

\begin{figure*}[!htbp]
	\centering{
		\includegraphics[width=2.0\columnwidth]{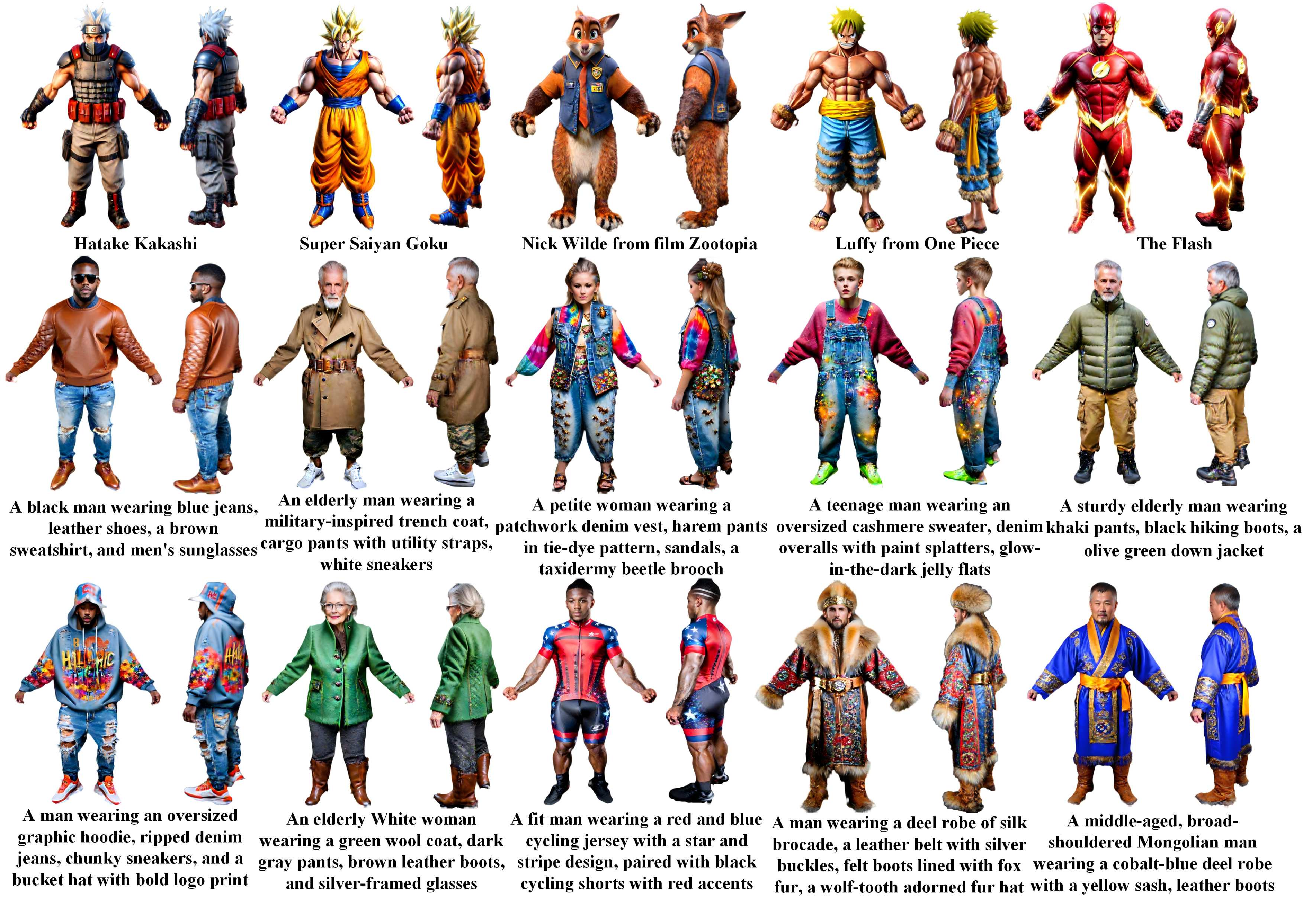}
		\caption{More examples of 3D huamn produced by our approach. The text prompts used are listed below.}
		\label{fig6}
	}
\end{figure*}

\noindent \textbf{Comparison with text-to-3D content methods.}
Furthermore, to verify the generalization capability of our method, we conduct a qualitative texture comparison with text-to-3D avatar generation approaches. We employ several commonly used benchmark cases for text-to-3D avatar synthesis. As illustrated in Fig. \ref{fig5}, DreamGaussian \cite{dreamgaussian} represents 3D content using 3DGS; however, due to the absence of avatar-specific priors, it suffers from severe model collapse and blurry textures. DreamAvatar \cite{dreamavatar} adopts SMPL-X as a geometric prior to ensure structural consistency, yet the generated textures remain overly smooth and lack fine-grained details. AvatarVerse \cite{avatarverse} employs a progressive generation strategy, improving overall texture fidelity but still failing to capture intricate surface details. In contrast, our method preserves the highest possible level of texture detail while maintaining geometric consistency. Consequently, our approach demonstrates superior robustness in texture refinement and detail preservation.

In summary, as illustrated in Fig. \ref{fig6}, we present additional cases to further demonstrate the effectiveness of our method. These include 10 complex long-text descriptions and 5 standard avatar generation cases. The experimental results validate the robustness of our approach, showcasing its ability to consistently generate high-fidelity 3D human models across diverse input prompts and varying levels of textual complexity.

\subsection{Quantitative Evaluation}

\noindent \textbf{User Study}. To ensure a fair and comprehensive comparison with existing methods, we conducted a user study evaluating the quality of generated 3D human models. Specifically, we randomly selected 100 cases from a set of 50 complex long-text descriptions. A total of 20 participants were recruited to provide subjective evaluations. Each participant reviewed the rendered 3D human models produced by different methods and rated them based on two key criteria: texture quality and geometric fidelity. The ratings were assigned on a scale from 1 to 10, with higher scores indicating superior performance. The evaluation results are presented in Table \ref{tab}.

Notably, our method consistently outperforms all competing approaches across both evaluation metrics. In particular, our approach excels in texture quality, precisely capturing fine-grained geometric structures and realistic details. These findings highlight the robustness and effectiveness of our method in generating high-fidelity 3D human models that closely align with user expectations and textual descriptions. Furthermore, this user study underscores the practical advantages of our approach, demonstrating its ability to maintain exceptional visual realism and semantic consistency even in complex multi-attribute scenarios.
\begin{table*}[htbp]
	\caption{Quantitative comparisons with text-to-3D methods.}
	\begin{center}
		\begin{tabular}{cccccc}
			\toprule
			\textbf{Method} & \textbf{ Texture Quality (↑)} & \textbf{Geometry Quality (↑)} &\textbf{CLIP Score (↑)} & \textbf{DenseScore Score (↑)} &\textbf{HPS\_V2 Score (↑)} \\
			\midrule
			DreamWaltz &5.4& 4.0 & 28.53& 24.75&24.31 \\
			TADA & 6.5 & 7.4 &29.16 &24.93 & 25.47 \\
			X-Oscar &7.3& 7.6 &30.54 &25.24 & 25.84 \\
			HumanGaussian & 8.3 & 7.6&30.67 &29.68 &30.76  \\
			HumanNorm & 8.2 & 7.9 &30.48 &29.72 & 30.64 \\
			\midrule
			Ours & \textbf{8.4} & \textbf{8.0}  &\textbf{32.09} &\textbf{31.47} & \textbf{32.51}\\
			\bottomrule
		\end{tabular}
		\label{tab}
	\end{center}
\end{table*}

As shown in Table \ref{tab}, DenseScore \cite{visionprefer} imposes more stringent requirements, leading to relatively lower scores for other models compared to other evaluation metrics. However, our method achieves an outstanding DenseScore, indicating that the multi-view images generated from our 3D human models exhibit superior semantic consistency. Additionally, the high CLIP score demonstrates that the generated views align closely with the input textual descriptions. Furthermore, the high HPS\_V2 \cite{hps} score suggests that our approach preserves richer texture details and exhibits a higher level of aesthetic quality. These results underscore the effectiveness of our method in generating 3D human models that simultaneously achieve high aesthetic fidelity and semantic accuracy.

\subsection{Ablation Study}

\begin{figure*}[!htbp]
	\centering{
		\includegraphics[width=2.0\columnwidth]{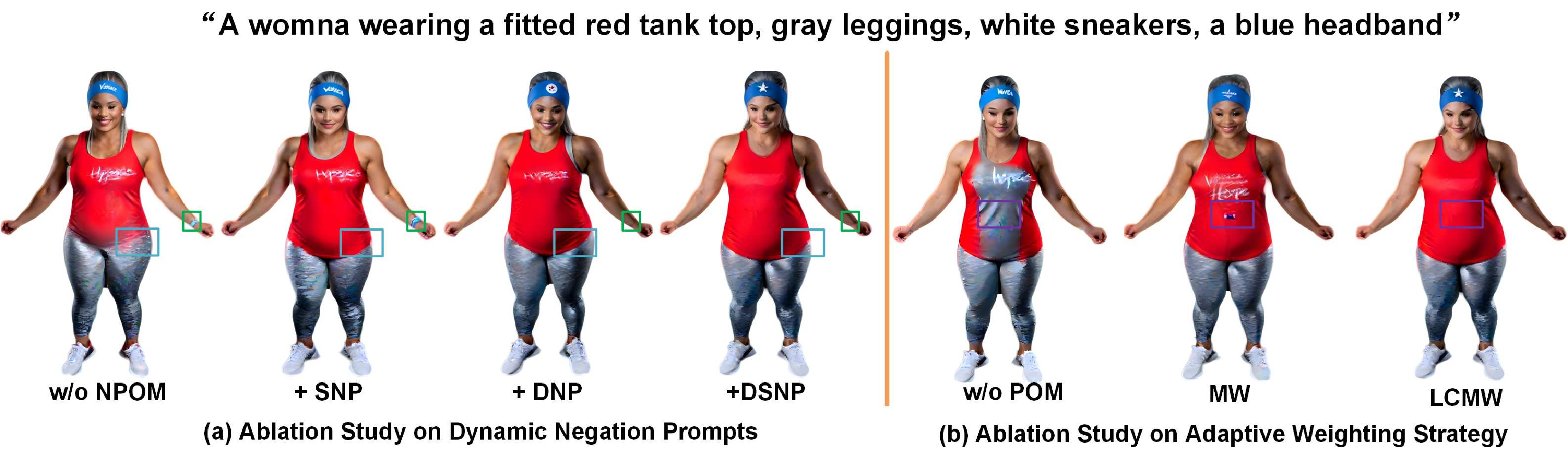}
		\caption{Ablation Study on Negation Prompts and Adaptive Weighting Strategy
			The proposed method is evaluated under the following conditions:
			(a) w/o NPOM: Experiments are conducted by incrementally incorporating SNP, DNP, and DSNP (DNP + SNP).
			(b) w/o POM: The absence of POM leads to optimization bias, resulting in deviations in generation quality. While w/ NPOM helps eliminate irrelevant objects, proper preference balancing remains crucial.
			Furthermore, we compare two weighting strategies: LCMW and MW to balance the contributions of different preference models. The results indicate that LCMW achieves superior optimization performance, leading to better text-to-3D alignment and enhanced generation quality.}
		\label{fig7}
	}
\end{figure*}
\begin{figure}[!htbp]
	\centering{
		\includegraphics[width=1.0\columnwidth]{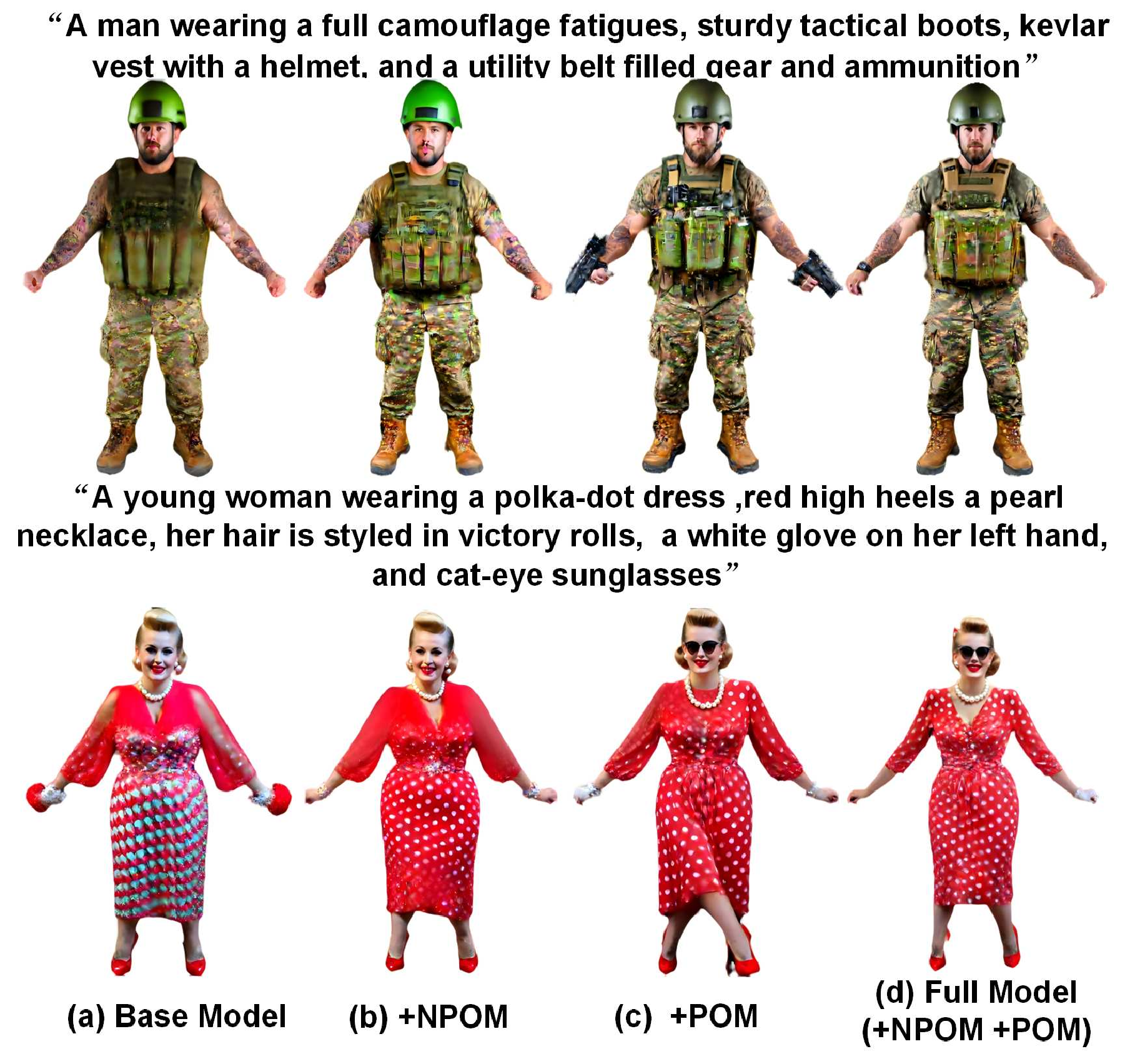}
		\caption{Ablation study of our method. We present the results of frontal human view generation under two ablation settings to better visualize and compare: (a) Base model; (b) +NPOM; (c) +POM; and (d) Full model.}
		\label{fig4}
	}
\end{figure}

\noindent \textbf{Effect of Preference Optimization Module.}
In Fig. \ref{fig4}(c), we introduce the Preference Optimization Model (POM) to evaluate its impact on enhancing the geometric and texture consistency of 3D human generation under complex long-text descriptions. Experimental results demonstrate that POM significantly improves generation quality, particularly in handling intricate prompts. For instance, in the first example, the generated 3D human successfully restores the camouflage jacket’s texture details and accurately depicts the multifunctional belt equipped with gear and ammunition, achieving high fidelity to the textual description. In the second example, the model precisely generates fine-grained visual elements, such as cat-eye glasses and gloves, further validating POM’s advantage in modeling complex visual attributes.

Table \ref{tab1} further quantifies the contribution of POM to generation quality. When POM is removed (i.e., using only NPOM), both CLIP and HPS\_V2 scores drop significantly, highlighting the crucial role of POM in enhancing text-image alignment.
Furthermore, to verify the effectiveness of the adaptive weighting strategy (LCMW), we compare it with mean weighting (MW). As illustrated in Fig. \ref{fig7}-(b), MW fails to balance the contributions of different preference models, leading to optimization failure and the unintended generation of artifacts on the tank top. In contrast, Table. \ref{tab1} shows that LCMW achieves higher scores, dynamically adjusting the contributions of different preference models to improve overall performance.
This observation further confirms the superiority of LCMW in optimizing 3D human generation, ensuring robust preference integration and enhanced semantic alignment.

\noindent \textbf{Effect of Negative Preference Optimization Module.}
As shown in Fig. \ref{fig4} (b)–(d), the incorporation of the POM may lead to over-optimization, resulting in unintended artifacts. For instance, in the first example, an extraneous firearm appears despite not being mentioned in the input description, while in the second case, an undescribed glove is erroneously generated on the right hand. To alleviate this issue, we introduce Dynamic Negation Prompts (DNP), which dynamically identify unintended objects, such as firearms and mirrored gloves, and employ the Negative Preference Optimization Model (NPOM) to suppress these artifacts effectively. The quantitative results in Table \ref{tab1} further demonstrate the necessity of NPOM, as removing it (i.e., using only POM with LLMW) leads to a notable decline in DenseScore, highlighting its effectiveness in refining text-to-3D generation.

To systematically evaluate the effectiveness of negative prompts, we compare three configurations: NPOM with Static Negation Prompts (SNP), NPOM with Dynamic Negation Prompts (DNP), and NPOM with a hybrid Dynamic-Static Negation Strategy (DSNP).

As shown in Fig. \ref{fig7}-(a), the absence of NPOM (w/o NPOM) leads to over-optimization, resulting in the spurious generation of a wristband and fusion between the tank top and leggings. Incorporating SNP mitigates the fusion artifact but fails to eliminate the unintended wristband. In contrast, DNP leverages LLM-driven estimation of potential irrelevant elements and systematic reconstruction of apparel attributes, effectively suppressing both the wristband artifact and clothing fusion issues. Furthermore, DSNP, which combines static and dynamic negation strategies, further enhances texture details and preserves a more realistic 3D human representation.

As presented in Table \ref{tab1}, DNP consistently outperforms SNP in DenseCLIP evaluations, demonstrating its superior capability in suppressing irrelevant artifacts while maintaining strong semantic alignment with textual descriptions. Overall, NPOM effectively alleviates the over-optimization issues introduced by POM, ensuring the generation of high-fidelity, artifact-free 3D human models with improved visual realism and semantic consistency.

\begin{table}[htbp]
	\caption{Quantitative Comparison of Ablation Components.}
	\begin{center}
		\begin{tabular}{cccc}
			\toprule
			\textbf{Method} & CLIP Score & DenseScore Score & HPS\_V2 Score\\
			\midrule
			w/o POM      & 30.15   &28.84  & 29.57 \\		
			w/ POM (MW)  & 30.76   & 28.41 & 30.94 \\
			w/ POM (LCMW) &31.54  &29.27   &31.16  \\
			\midrule
			w/o NPOM      & 30.70  & 29.47 &29.63  \\
			w/ NPOM(SNP)  & 30.78  & 29.58 &30.06 \\
			w/ NPOM(DNP)  & 30.62  &29.74  &29.67 \\
			w/ NPOM(DSNP) &30.78  &29.97  &30.12 \\                 
			\midrule   
			Ours & \textbf{32.09} & \textbf{31.47}  & \textbf{32.51} \\
			\bottomrule
		\end{tabular}
		\label{tab1}
	\end{center}
\end{table}

\begin{figure}[!htbp]
	\centering{
		\includegraphics[width=1.0\columnwidth]{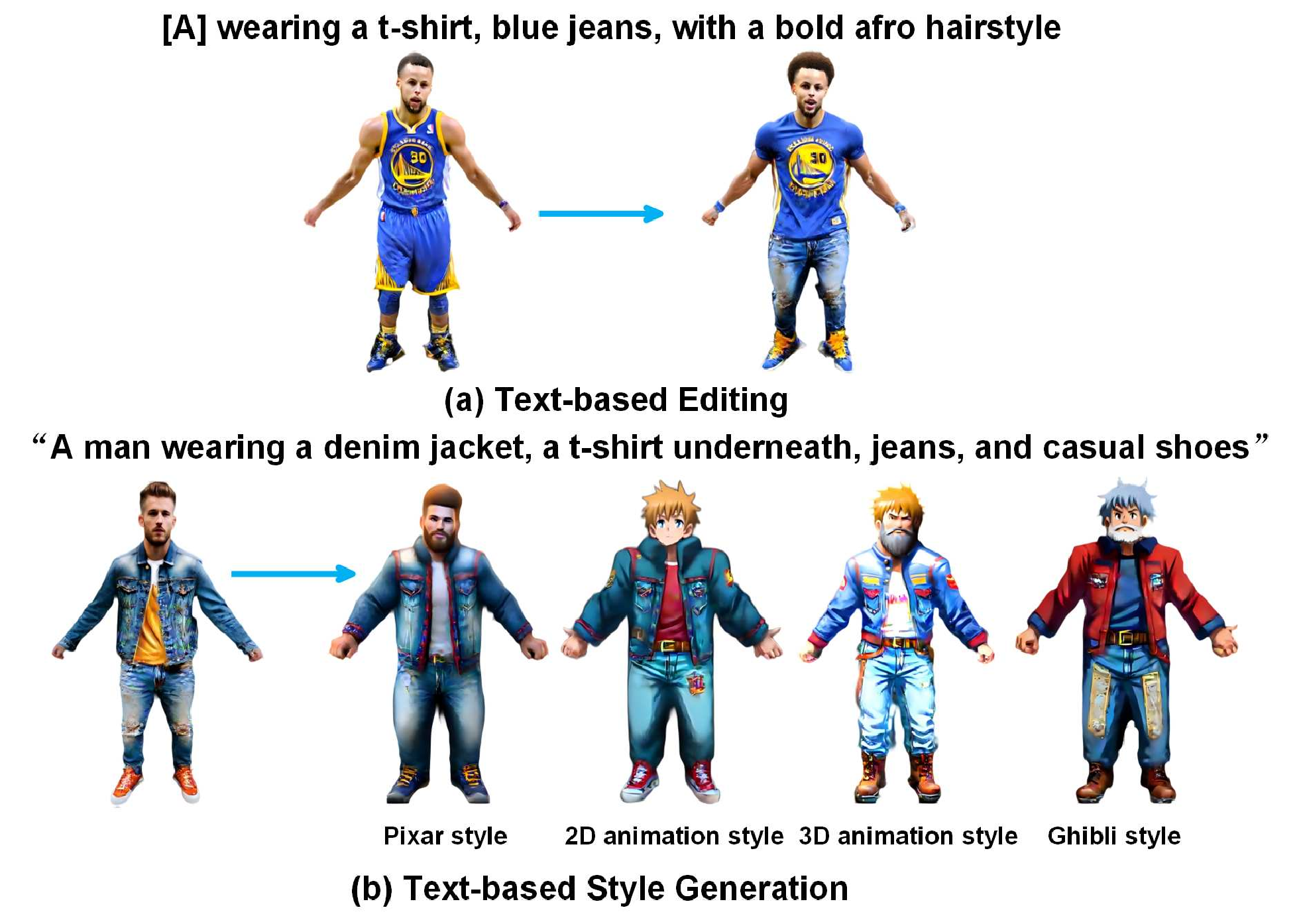}
		\caption{Applications of our method.}
		\label{fig2}
	}
\end{figure}

\subsection{Applications}

\noindent \textbf{Text-based Editing.}
Our method enables precise texture editing of 3D human models by adjusting textual prompts, allowing control over attributes such as clothing color, style, and hairstyle. As shown in Fig. \ref{fig2} (a), we demonstrate this capability by modifying Curry’s appearance. This flexibility ensures high visual fidelity and personalization, benefiting applications in content creation, virtual avatars, and digital entertainment.

\noindent \textbf{Text-based Style Creation.}
Our method enables the generation of 3D human models in diverse styles by incorporating specific style types into textual prompts, allowing precise control over attributes like color schemes and visual aesthetics. As shown in Fig. \ref{fig2} (b), we demonstrate this by modifying the man’s style and color. This adaptability enhances customization, making it particularly useful for virtual avatars, fashion design, and digital content creation.

\section{Conclusion}
We propose a contrastive preference modeling framework to enhance text-driven 3D human generation, improving semantic alignment and visual fidelity in SDS. By integrating positive and negative prompts, our method refines text-conditioned generation, ensuring structural coherence and artifact suppression.
Key contributions include a preference optimization module for better semantic understanding and a negation preference module leveraging static-dynamic negation prompts to eliminate inconsistencies. Additionally, our LLM-driven dynamic negation prompting further mitigates artifacts and prevents reward hacking.
Extensive experiments demonstrate state-of-the-art performance in texture realism and semantic alignment, particularly for complex textual descriptions. Future work will explore extending this approach to broader 3D generation tasks.

\noindent \textbf{Limitations and future work.} Despite promising results, challenges like texture resolution limitations and Gaussian artifacts persist. Future work will focus on refining SDS to address oversaturation and blurriness, enhancing the accuracy and stability of text-driven 3D human generation.

\end{document}